\documentclass{article}

\PassOptionsToPackage{numbers, compress}{natbib}

\usepackage[final]{nips_2016}

\usepackage[utf8]{inputenc} 
\usepackage[T1]{fontenc}    
\usepackage{hyperref}       
\usepackage{url}            
\usepackage{booktabs}       
\usepackage{amsfonts}       
\usepackage{nicefrac}       
\usepackage{microtype}      
\usepackage{comment}
\usepackage{graphicx}

\title{Towards the Creation of a Large Corpus of Synthetically-Identified Clinical Notes}

\author{
  Willie Boag\\
  MIT CSAIL \\
  Cambridge, MA, 02139 \\
  \texttt{wboag@mit.edu} \\
  \And
  Tristan Naumann\\
  MIT CSAIL \\
  Cambridge, MA, 02139 \\
  \texttt{tjn@mit.edu} \\
  \And
  Peter Szolovits\\
  MIT CSAIL \\
  Cambridge, MA, 02139 \\
  \texttt{psz@mit.edu} \\
}

\begin{document}

\maketitle

\begin{abstract}
Clinical notes often describe the most important aspects of a patient's physiology and are therefore critical to medical research. However, these notes are typically inaccessible to researchers without prior removal of sensitive protected health information (PHI), a natural language processing (NLP) task referred to as de-identification. Tools to automatically de-identify clinical notes are needed but are difficult to create without access to those very same notes containing PHI. This work presents a first step toward creating a large synthetically-identified corpus of clinical notes and corresponding PHI annotations in order to facilitate the development de-identification tools. Further, one such tool is evaluated against this corpus in order to understand the advantages and shortcomings of this approach.
\end{abstract}

\section{Introduction}
\label{sec:Introduction}
In natural language processing (NLP), access to data is critical in the development of high-quality tools. Indeed, \citet{halevy2009unreasonable} remarked on the ``unreasonable effectiveness of data'' insofar as simple models using sufficient data often outperform more complex or clever models using less data. This realization presents a unique challenge within the clinical domain.

Electronic Health Record (EHR) notes are an important source of information for improving our current understanding of patients~\citep{ghassemi2014unfolding}. However, text typically contains sensitive protected health information (PHI) and consequently cannot be shared easily among researchers due to legal limitations established to ensure patient privacy. These restrictions have led to NLP systems in the clinical domain that often lag behind those in the general domain, even for tasks that are essentially considered solved~\citep{boag2015cliner}. Improving clinical NLP, while retaining patient privacy, requires de-identified clinical notes, those from which PHI has been redacted.

Manual approaches to de-identification are unsuitable as modern EHRs are rapidly growing. Automatic solutions benefit from scalability, but existing systems are hamstrung by a set of mutually dependent circumstances: developing de-identification systems requires access to notes, but these very same notes cannot be made available prior to de-identification. With this in mind \citet{uzuner2007evaluating} created the 2006 Informatics for Integrating Biology and the Bedside (i2b2) challenge for automatic de-identification of clinical records. The shared task oversaw the creation of a corpus of $889$ discharge summaries in which PHI was annotated, removed, and replaced with out-of-vocabulary PHI surrogates. 
The 2014 i2b2 challenge revisited the task of de-identification, providing the largest known corpus of $1304$ longitudinal records for $296$ patients~\citep{stubbs2015annotating}.

While exceptional in quality, the i2b2 corpora together provide just over $2000$ records. \citet{ferrandez2012evaluating} note that access to sufficient data for machine learning approaches stands to provide improved precision relative to prior systems based on pattern matching and rules. Indeed, recent work by \citet{dernoncourt2016identification} demonstrates the potential of systems with access to substantially more data. Using this motivation, the present work makes the following two contributions:
\begin{enumerate}
\item Provide first steps toward a corpus of clinical notes that have been synthetically-identified, meaning they have been de-identified and PHI has been replaced with reasonable surrogates.
\item Evaluate a baseline de-identification system against this data in order to understand advantages and shortcomings of this approach.
\end{enumerate}

\section{Methods}
\label{sec:Methods}

\subsection{Synthetic-Identification}
MIMIC-III is an publicly available dataset developed by the MIT Lab for Computational Physiology, containing data for $61,532$ ICU stays for $46,520$ patients~\citep{mimic3}. MIMIC-III contains over 2 million de-identified clinical notes with over 12.5 million instances of PHI among nearly 500 million tokens as shown in Table~\ref{tab:mimic3notes}. To make these notes publicly available, \citet{neamatullah2008automated} developed a rule-based system tailored to the notes that appear in MIMIC. 

Synthetically-identified notes were created by substituting reasonable surrogate values for annotated PHI. E.g., replacing ``[**Patient Name**] visited [**Hospital**]'' with ``Mary Smith visited MGH.'' Though patient identity coference was not performed for this initial work, we hope to incorporate it into the notes by reverse engineering the naming conventions from the original de-identification script. Names were chosen proportional to their frequency from 2005 U.S. Census~\footnote{Most Common First and Last Names in the U.S.: http://names.mongabay.com/} and hospitals were chosen uniformly from a list of U.S. hospitals~\footnote{List of Hospitals in the U.S.: en.wikipedia.org/wiki/Lists\_of\_hospitals\_in\_the\_United\_States}.
While many types of PHI exist in MIMIC, patient names and hospitals are abundant; therefore, initial efforts have been directed toward these two categories -- the focus of our subsequent experiments. 

\begin{table}[t]
  \caption{Distribution of de-identified protected health information (PHI) in MIMIC-III v1.4 notes by category. Particularly noteworthy are the \~12.5 million instances of PHI among \~500 million tokens.}
  \label{tab:mimic3notes}
  \centering
  \begin{tabular}{r|llr|llr}
    \toprule
    Category & Notes & \multicolumn{2}{c}{Contain PHI} & Tokens & \multicolumn{2}{c}{PHI Instances} \\
    \midrule
    Case Management   &    967 &    954 & (98.65\%) &    131806 &    9860 & (7.48\%)\\
	Consult           &     98 &     98 &   (100\%) &     71453 &    1843 & (2.58\%)\\
	Discharge summary &  59652 &  59651 & (99.99\%) &  80986971 & 2632527 & (3.25\%)\\
	ECG               & 209051 & 133146 & (63.69\%) &   5856486 &  135048 & (2.31\%)\\
	Echo              &  45794 &  45794 &   (100\%) &  14817189 &  127233 & (0.86\%)\\
	General           &   8301 &   5200 & (62.64\%) &   1688905 &   36923 & (2.19\%)\\
	Nursing           & 223556 & 188691 & (84.40\%) &  56107626 & 1048996 & (1.87\%)\\
	Nursing/other     & 822497 & 561187 & (68.23\%) & 104063367 & 1718441 & (1.65\%)\\
	Nutrition         &   9418 &   9196 & (97.64\%) &   3068351 &  204730 & (6.67\%)\\
	Pharmacy          &    103 &     96 & (93.20\%) &     34466 &    1100 & (3.19\%)\\
	Physician         & 141624 & 141047 & (99.59\%) & 115484159 & 3475738 & (3.01\%)\\
	Radiology         & 522279 & 522278 & (99.99\%) & 102460089 & 3097379 & (3.02\%)\\
	Rehab Services    &   5431 &   5010 & (92.24\%) &   2125724 &   52955 & (2.49\%)\\
	Respiratory       &  31739 &  10395 & (32.75\%) &   4717416 &   14662 & (0.31\%)\\
	Social Work       &   2670 &   2609 & (97.72\%) &    779550 &   38691 & (4.96\%)\\
    \midrule
    {\bf Total} & 2083180 & 1685352 & (80.90\%) & 492393558 & 12596126 & (2.56\%) \\
    \bottomrule
  \end{tabular}
\end{table}

\subsection{De-Identification}
De-identification was performed with a Conditional Random Field (CRF), a statistical modeling method which has proven to be effective for the de-identification problem~\citep{ferrandez2012evaluating,uzuner2007evaluating}. The system was trained to identify a variety of PHI tags, including: names, hospitals, locations, dates, and identifying numbers (e.g. SSN). 

For each word, $w_i$, the CRF model made use of the following features:
\begin{enumerate}
    \item Presence/Absence of each word in the training vocabulary (one hot where only $w_i$ is on)
    \item Presence/Absence of previous three words (only $w_{i-1}$, $w_{i-2}$, and $w_{i-3}$ are on)    
    \item Presence/Absence of next three words (only $w_{i+1}$, $w_{i+2}$, and $w_{i+3}$ are on)
    \item Whether $w_i$ occurred in lists of common male names, female names, surnames, or hospitals (4 features in total)
\end{enumerate}

Rather than padding the beginning and end of sentences with special START/STOP symbols for those features which inspect previous or next tokens, the features were instead generated by reading from adjacent lines in the note (i.e., only the beginning and end of each note was padded rather than each statement). This was done because the free-form structure of the various notes--including alternating prose and nonprose unpredictably, insertion of line-breaks midway through prose, and variety of note types--are difficult to parse without special consideration. 
Instead, we provide the model with the ability to learn these relationships automatically.

\section{Results}
\label{sec:Results}
Evaluation was performed at the token-level, so that multi-token PHI instances are treated separately during evaluation. The motivation for this is that it gives a good sense for what percentage of PHI is being detected: {\it exact span} metrics cannot differentiate between a decent method that gets most-but-not-all tokens in a span and {\it inexact span} metrics might give too much partial credit to systems that are not identifying all relevant PHI. We felt that token-level precision, recall, and F1 provided the most interpretable and reasonable metric.

\begin{figure}
    \centering
    \caption{The effect that the number of training files has on HOSPITAL recall.}
    \includegraphics[width=70mm]{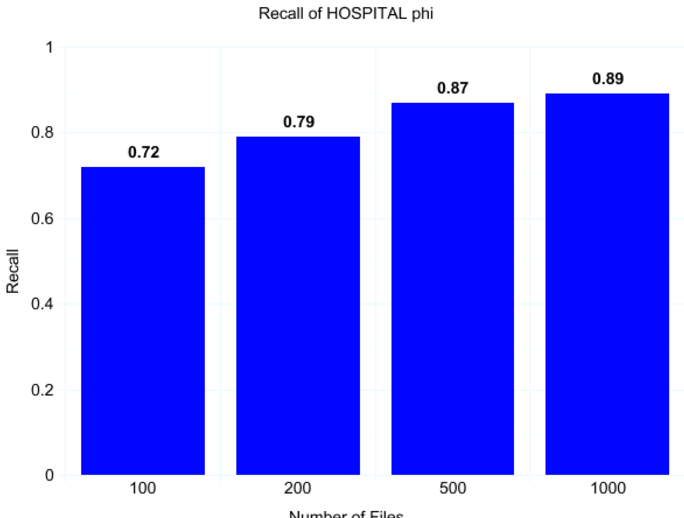}
    \label{fig:n-patient-recall}
\end{figure}

Unsurprisingly, our experiments show a simple trend: the more data, the better. We can see from Table~\ref{tab:results-all} that as we increase the number of files used for training, we see improvements in all scores for both patient name and hospital retrieval. In particular, we can see that although precision is already very good with a small training data, the recall continues to show sizable improvements as we train on more data, as indicated by Figure~\ref{fig:n-patient-recall}.
These results demonstrate the benefits that a large-scale dataset can offer, especially for complex de-identification models involving deep learning approaches.

\begin{table}[ht]
\caption{The results for patient name and hospital PHI identification as we add more training data.}
\begin{center}
\begin{tabular}{ccc}
	\footnotesize
	\begin{minipage}{.5\linewidth}
		\begin{tabular}{|l|l|l|l|}
			\hline
			\multicolumn{4}{|c|}{PATIENT\_NAME}\\ \hline
			\# training files          &  precision & recall & f1 \\ \hline
		    100        &  0.89      &  0.78  & 0.83  \\ \hline
   		 	200        &  0.90      &  0.87  & 0.88  \\ \hline
   		 	500        &  0.91      &  0.93  & 0.92  \\ \hline
 		   	1000       &  0.92      &  0.96  & 0.94  \\ \hline
            \hline
		\end{tabular}
    \end{minipage} &
    
    \footnotesize
	\begin{minipage}{.45\linewidth}
		\begin{tabular}{|l|l|l|l|}
			\hline
			\multicolumn{4}{|c|}{HOSPITAL}\\ \hline
			\# training files &  precision & recall & f1 \\ \hline
		    100        &  0.92      &  0.72  & 0.81  \\ \hline
   		 	200        &  0.93      &  0.79  & 0.86  \\ \hline
   		 	500        &  0.94      &  0.87  & 0.91  \\ \hline
 		   	1000       &  0.95      &  0.89  & 0.92  \\ \hline
            \hline
		\end{tabular}
    \end{minipage} &
\end{tabular}
\end{center}
\label{tab:results-all}
\end{table}

\section{Conclusions}
\label{sec:Conclusions}
While this work represents an important first step toward the creation of a large corpus of synthetically-identified clinical notes, it suffers from several limitations. Notably, \citet{dernoncourt2016identification} note that the system created by \citet{neamatullah2008automated} favors recall over precision as it introduces virtually no false negatives, while there are numerous false positives accounting for up to 15\% of the PHI instances detected. This suggests that creation of the synthetically-identified dataset directly from the identified data is necessary so as not to unnecessarily introduce meaningless surrogate data. 


Further, we recognize that the using surrogate data may constrain the natural variety of names, places, and other entities that appear in notes (e.g., through misspellings); thus, making the task slightly easier. Again, it is the hope that with such a large volume of PHI annotations this will become decreasingly important relative to the potential improvement in tools possible.

Nevertheless, this work marks a first attempt to providing the community with a synthetically-identified corpus of notes. The authors hope that this may be subsequently improved through collaboration with the MIT Laboratory of Computational Physiology to provide a higher-quality version of synthetically-identified notes. While the issue of co-reference makes this problem more difficult that the usual task of de-identification (which most closely resembles named entity recognition), providing synthetically-identified notes makes the identification of errors less conspicuous to the human reader. It may therefore be a preferable means of distributing notes as well.

\subsubsection*{Acknowledgments}
This research was funded in part by the Intel Science and Technology Center for Big Data, the National Library of Medicine Biomedical Informatics Research Training grant (NIH/NLM 2T15 LM007092-22).

This material is based upon work supported by the National Science Foundation Graduate Research Fellowship Program under Grant No. 1122374. Any opinions, findings, and conclusions or recommendations expressed in this material are those of the author(s) and do not necessarily reflect the views of the National Science Foundation.

\bibliographystyle{abbrvnat}
\bibliography{ml4hc}
\end{document}